%% file: arxiv.tex
\documentclass{article}
\usepackage[T1]{fontenc}
\usepackage{longcat_style,times}
\usepackage{natbib}
\input{datapraxis_title}
\usepackage{amsmath,amssymb,graphicx,booktabs,array,multirow}
\usepackage{hyperref}
\usepackage{xurl}
\usepackage{wrapfig,placeins,float,capt-of}
\usepackage{colortbl}
\usepackage{multirow}
\usepackage{framed,needspace}

\definecolor{datapraxisblue}{RGB}{235,244,252}
\hypersetup{
  hidelinks,
  pdftitle={DataPraxis: Scaling Analytics Tasks via Verified Workspace Evolution},
  pdfauthor={Nan Jia, Wenxuan Shi, Jingqing Ruan, Zhizhao Zeng, Tianyi Zhai, Zhongda Su, Feng Hong, Zhengyu Chen, Wenhao Yan, Haojin Yang, Siqi Dai, Weipeng Zhang, Ke Zeng, Xunliang Cai}
}
\title{DataPraxis: Scaling Analytics Tasks via \\Verified Workspace Evolution}
\input{authors_arxiv}
\begin{document}
\maketitle
\begin{abstract}
\input{sections/abstract}
\end{abstract}
\input{sections/introduction}
\input{sections/method}

\input{sections/experiments_v2}

\input{sections/related_work}
\section{Conclusion}
\label{sec:conclusion}
We introduced DataPraxis, a framework that uses verified workspace evolution to synthesize professional analytical tasks.
By accumulating and combining analytical findings, it uncovers deeper questions that are difficult to formulate from raw materials alone.
Verification draws on evolution evidence alongside solution evidence to establish task criteria and filter training trajectories.
Experiments with DataPraxis-5K demonstrated improved agent performance across two model scales and multiple benchmarks.
These findings support the value of accumulated workspace knowledge as a foundation for synthesizing deeper tasks.
DataPraxis offers a promising approach to scaling task difficulty, opening new possibilities for pushing the boundaries of model capabilities.

\input{sections/limitations}
\input{sections/ai_use_statement}

\bibliography{references}
\bibliographystyle{iclr2026_conference}
\clearpage
\appendix
\input{sections/appendix_ridebench}

\input{sections/appendix_case_studies}

\input{sections/appendix_agent_prompts}
\end{document}

%% file: datapraxis_title.tex
\renewcommand{\headeright}{\raisebox{-0.2\height}{\includegraphics[height=2.2em]{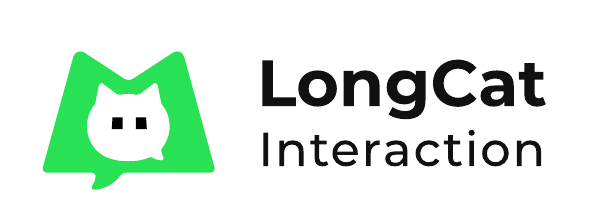}}}
\renewcommand{\shorttitle}{DataPraxis}
\makeatletter
\renewcommand{\@maketitle}{%
  \vbox{%
    \hsize\textwidth
    \linewidth\hsize
    \vskip -0.5in
    \noindent
    \begin{minipage}{0.2\textwidth}
      \includegraphics[width=\linewidth]{pic/Logo.pdf}
    \end{minipage}%
    \\
    \rule{\linewidth}{1pt}
    \hspace{0.05\textwidth}%
    \begin{minipage}{0.8\textwidth}
    \end{minipage}

    \centering
    {\LARGE \bfseries\@title\par}
    \vskip 0.1in  
    \def\And{%
      \end{tabular}\hfil\linebreak[0]\hfil%
      \begin{tabular}[t]{c}\bf\rule{\z@}{24\p@}\ignorespaces%
    }
    \def\AND{%
      \end{tabular}\hfil\linebreak[4]\hfil%
      \begin{tabular}[t]{c}\bf\rule{\z@}{24\p@}\ignorespaces%
    }
    \begin{tabular}[t]{c}\bf\rule{\z@}{24\p@}\@author\end{tabular}%
  \vskip 0.05in 
  }
}
\makeatother

%% file: authors_arxiv.tex
\author{%
  Nan Jia\textsuperscript{1,}\thanks{\raggedright Equal contribution. Emails: \href{mailto:jianan@hust.edu.cn}{\texttt{jianan@hust.edu.cn}}, \href{mailto:shiwenxuan20@mails.ucas.ac.cn}{\texttt{shiwenxuan20@mails.ucas.ac.cn}}.}\textsuperscript{\phantom{*},\textdaggerdbl}, Wenxuan Shi\textsuperscript{2,*,\textdaggerdbl},
  Jingqing Ruan\textsuperscript{3,}\thanks{\raggedright Corresponding authors. Emails: \href{mailto:ruanjingqing@meituan.com}{\texttt{ruanjingqing@meituan.com}}, \href{mailto:zengzhizhao@meituan.com}{\texttt{zengzhizhao@meituan.com}}.}, Zhizhao Zeng\textsuperscript{3,\textdagger},
  Tianyi Zhai\textsuperscript{3}, Zhongda Su\textsuperscript{3}\\
  \textbf{Feng Hong\textsuperscript{3}, Zhengyu Chen\textsuperscript{3},
  Wenhao Yan\textsuperscript{4,}\thanks{\raggedright Work was done during internships at Meituan.}, Haojin Yang\textsuperscript{5,\textdaggerdbl},
  Siqi Dai\textsuperscript{2,\textdaggerdbl}, Weipeng Zhang\textsuperscript{3}}\\
  \textbf{Ke Zeng\textsuperscript{3}, Xunliang Cai\textsuperscript{3}}\\
  \textsuperscript{1}Huazhong University of Science and Technology\\
  \textsuperscript{2}University of Chinese Academy of Sciences\quad
  \textsuperscript{3}Meituan\\
  \textsuperscript{4}Tsinghua University\quad
  \textsuperscript{5}Peking University
}

%% file: sections/abstract.tex
General-purpose agents are increasingly expected to perform professional analytics work, which requires sustained reasoning, iterative investigation, and reliable evidence.
Training these capabilities requires diverse and challenging tasks. 
However, generating tasks directly from raw data often overlooks deeper questions that emerge through analysis and provides limited evidence for verification. 
To address these limitations, we introduce DataPraxis, an analytics task synthesis framework based on verified workspace evolution. 
The framework progressively enriches workspaces with validated findings and artifacts, then inherits and combines these workspaces to uncover deeper questions. 
It constructs tasks from accumulated findings and their dependencies, verifying analytical results both during evolution and after task assembly. 
Using this framework, we construct DataPraxis-5K, a dataset of 5,000 verified trajectories spanning diverse domains and output formats, to train DataPraxis-27B and DataPraxis-35B-A3B. 
DataPraxis-27B achieves 55.11 on DABStep and 51.9 on JobBench, matching or surpassing frontier models such as GLM-5.2.
DataPraxis-5K is available at \href{https://huggingface.co/datasets/stargazerhust/DataPraxis-5k}{HuggingFace}.

%% file: sections/introduction.tex
\section{Introduction}
\label{sec:introduction}

As large language models acquire capabilities for code execution, tool use, and multi-turn interaction, general-purpose agents are beginning to undertake professional work in real environments~\citep{yao2023react,wang2025openhands}. 
Analysis is a representative example: users expect models not only to answer questions, but also to understand the materials in a working directory, conduct analyses autonomously, and deliver well-supported conclusions and files~\citep{hong2024datainterpreter,zhang2025deepanalyze}. 
CLI agents such as Claude Code~\citep{anthropicClaudeCode}, Codex~\citep{openaiCodex}, and Pi Agent~\citep{piCodingAgent} provide a general-purpose execution interface for such work. 
However, training models to reliably complete these long-horizon analytical tasks still requires training tasks that cover complex analytical processes and provide reliable evaluation signals. 
Constructing such tasks at scale is a key challenge in improving general-purpose agents' professional analysis capabilities.

Existing agent task synthesis methods broadly fall into two categories. 
The first generates and combines tasks based on data descriptions and predefined structures~\citep{qiao2026datamind,xie2025agentsynth,shi2025taskcraft}. 
These methods guide task generation and increase complexity, but may struggle to discover tasks that emerge only through actual analysis. 
The second constructs tasks from interaction experience and execution evidence obtained through environment exploration~\citep{mai2025cues,chen2026dive}. 
These methods ground tasks in interaction evidence, but offer limited support for discovering deeper tasks that require further computation on prior results or integration of findings across analytical branches. 
Discovering deeper questions that are not apparent from raw materials through continued analysis therefore remains a key challenge in professional data analysis task synthesis. 
Meanwhile, intermediate findings are not inherently reliable, and errors can propagate when these findings are reused. 
Even when local findings are confirmed, the complete tasks constructed from them may still lack reliable evaluation criteria. 
Ensuring both the trustworthy accumulation of analytical results and reliable evaluation of the resulting tasks is another key challenge.

\edef\discoverySavedIntextsep{\the\intextsep}
\setlength{\intextsep}{6pt}
\begin{wrapfigure}{r}{0.45\textwidth}
\centering
\vspace{-6pt}
\includegraphics[width=\linewidth]{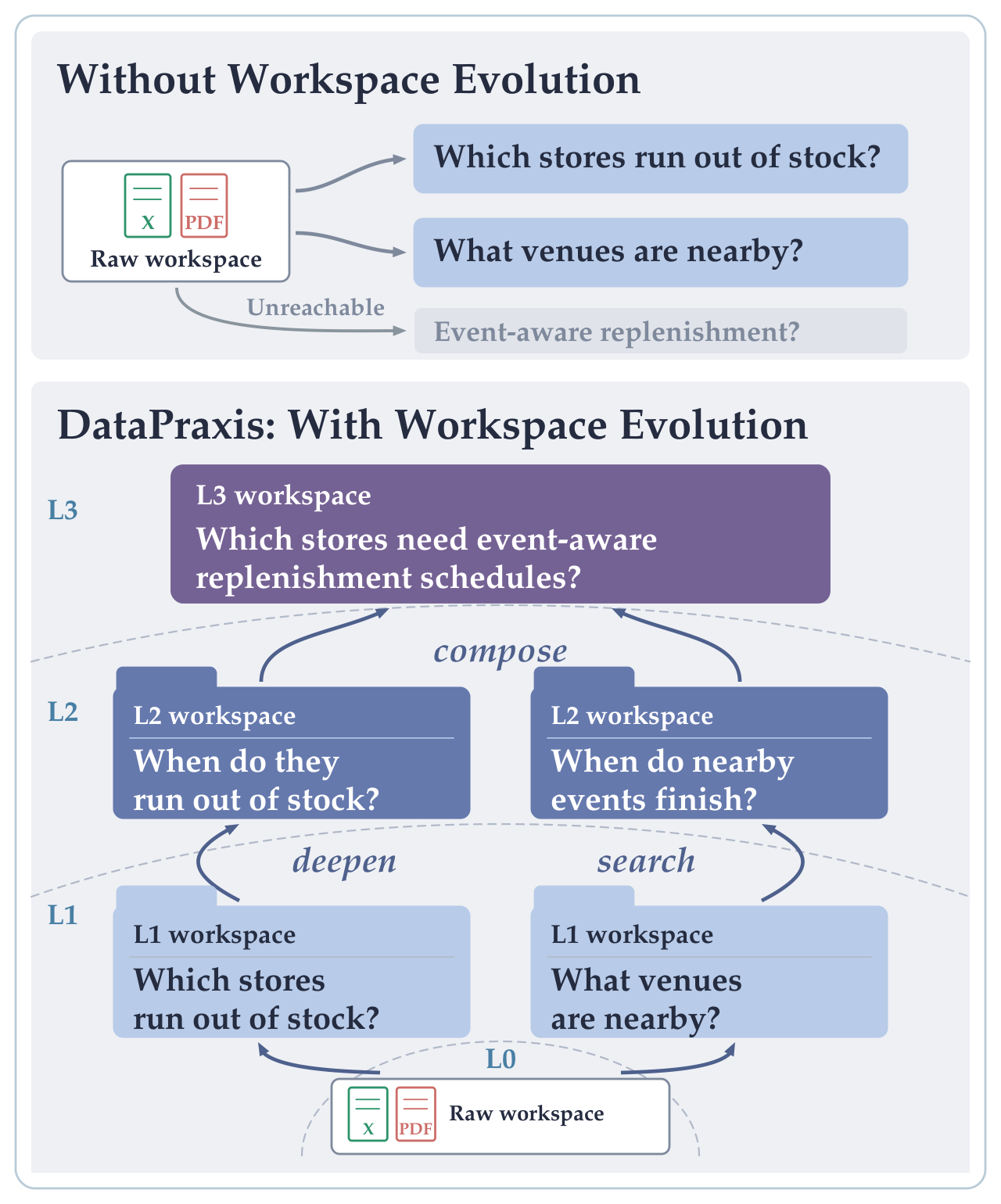}
\par\vspace{-6pt}
\setlength{\abovecaptionskip}{4pt}
\setlength{\belowcaptionskip}{0pt}
\caption[Workspace evolution for task discovery.]{\protect\raggedright \textbf{Up:} Tasks are generated from the initial materials without workspace evolution. \textbf{Down:} DataPraxis inherits and combines verified analytical results through workspace evolution, providing new grounds for discovering deeper questions.\protect\par}
\label{fig:discovery}
\end{wrapfigure}
\setlength{\intextsep}{\discoverySavedIntextsep}

To this end, we propose DataPraxis, a professional analytical task synthesis framework based on verified workspace evolution. 
Its central idea is that analytical results are not only answers to current questions, but also a basis for discovering subsequent ones. 
DataPraxis retains scripts, intermediate data, and analytical findings produced during execution in executable workspaces, enabling subsequent analysis to inherit prior results, explore new directions, and combine different branches for joint analysis. 
Figure~\ref{fig:discovery} illustrates the role of workspace evolution in task discovery: without continued workspace evolution, task generation cannot fully exploit progressively accumulated analytical results; by inheriting and combining verified workspaces, DataPraxis enriches the basis for proposing questions, progressively revealing deeper questions that are difficult to discover directly. 
The framework then constructs tasks backward from the resulting findings and their dependencies, removing intermediate support from the evolution process so that solvers must independently reconstruct the analysis from raw materials, thereby converting accumulated analytical dependencies into challenging long-horizon training tasks.
To ensure reliability from analytical discovery to task construction, DataPraxis introduces dual-evidence verification, combining verified findings from workspace evolution with independent solution trajectories to construct and validate evaluation rubrics. 
These rubrics assess final deliverables and filter solution trajectories for training.

Our main contributions are as follows:
\WFclear
\begin{itemize}
\item We propose DataPraxis, a data analysis task synthesis framework based on workspace evolution. 
Verified workspace evolution and dual-evidence verification enable the scalable construction of challenging, verifiable terminal-agent tasks.
\item We construct DataPraxis-5K, containing 5K long-horizon professional analytical training examples across 15 domains, including business, finance, sports, and entertainment. We train DataPraxis-27B, a 27B dense model, and DataPraxis-35B-A3B, a 35B mixture-of-experts model with 3B active parameters.
\item We demonstrate that workspace-evolution-based synthesis improves agents' ability to perform data analysis work. DataPraxis-27B achieves leading performance among the evaluated models on data analysis and long-horizon professional tasks.
\end{itemize}

%% file: sections/method.tex
\raggedbottom
\begin{figure}[t]
\centering
\includegraphics[width=0.95\linewidth]{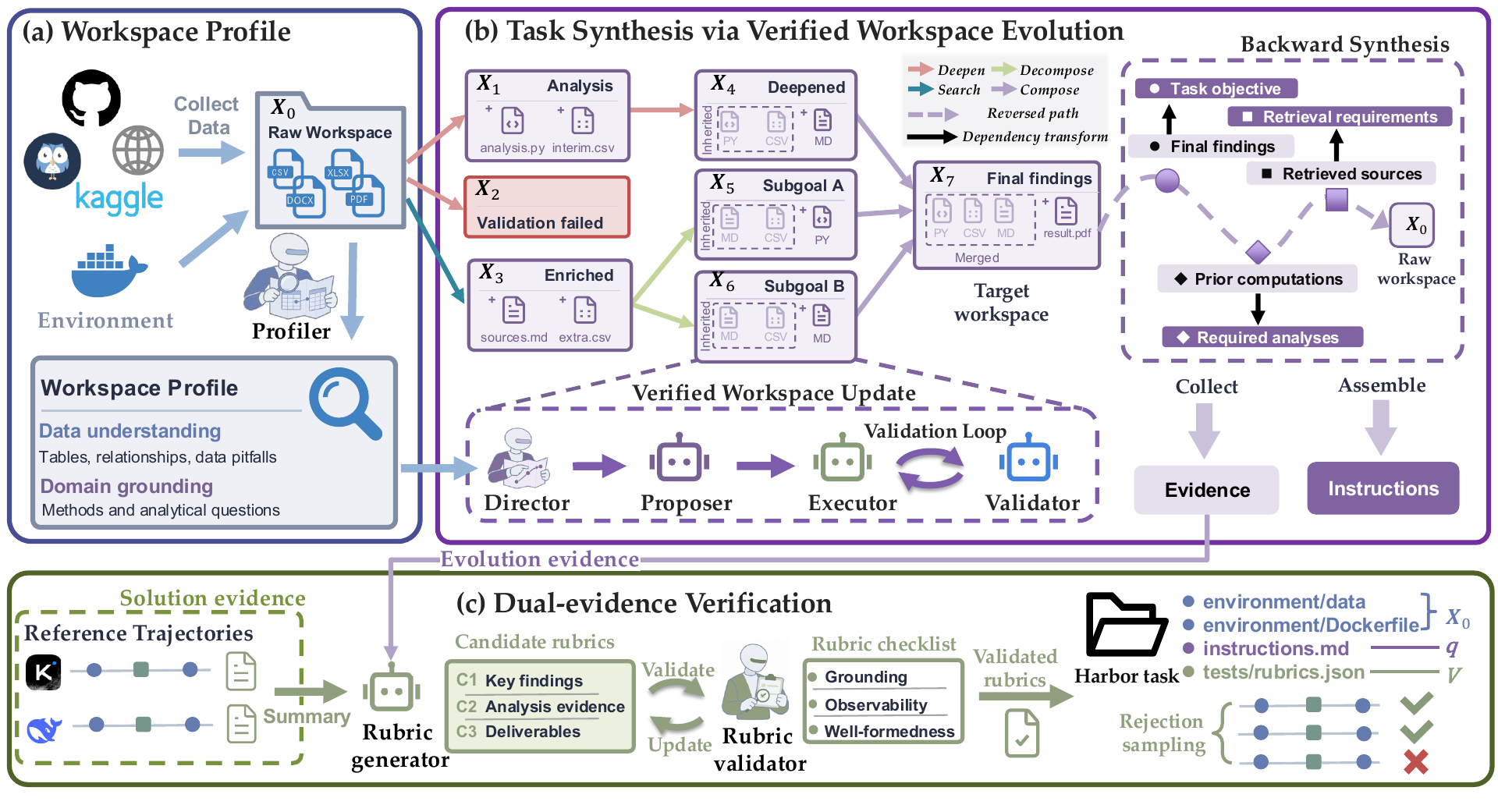}
\caption{The DataPraxis framework. 
(a) The Profiler characterizes the initial data and domain. 
(b) Verified workspace evolution progressively discovers and validates analytical findings, which are converted into standalone tasks through backward synthesis. 
(c) Dual-evidence verification combines evidence from evolution with independent solving trajectories.}
\label{fig:framework}
\end{figure}

\section{Motivation}
\label{sec:motivation}

Let $X_0=(\mathcal D,\mathcal E)$ denote a seed workspace containing data assets $\mathcal D$ and an execution environment $\mathcal E$.
DataPraxis constructs a task instruction $q$ and a verifier $V$ from this workspace:
\begin{equation}
X_0 \longrightarrow (q,V).
\label{eq:synthesis}
\end{equation}
The instruction $q$ specifies the analytical objective and delivery requirements for an agent operating in $X_0$. 
The verifier $V$ evaluates the analytical correctness, supporting evidence, and overall quality of the agent's final output.
The challenge is to formulate deep analytical questions from the initial workspace, where many patterns and relationships have yet to be
uncovered.
Findings uncovered during analysis can motivate questions that are not apparent from the initial workspace alone.

To examine how analysis informs subsequent questions, an agent investigates three successive questions in each of 50 workspaces, using findings from one analysis to formulate the next question.
\begin{wrapfigure}{R}{0.42\textwidth}
\centering
\includegraphics[width=0.90\linewidth,trim={36.6bp 14.7bp 15.9bp 15bp},clip]{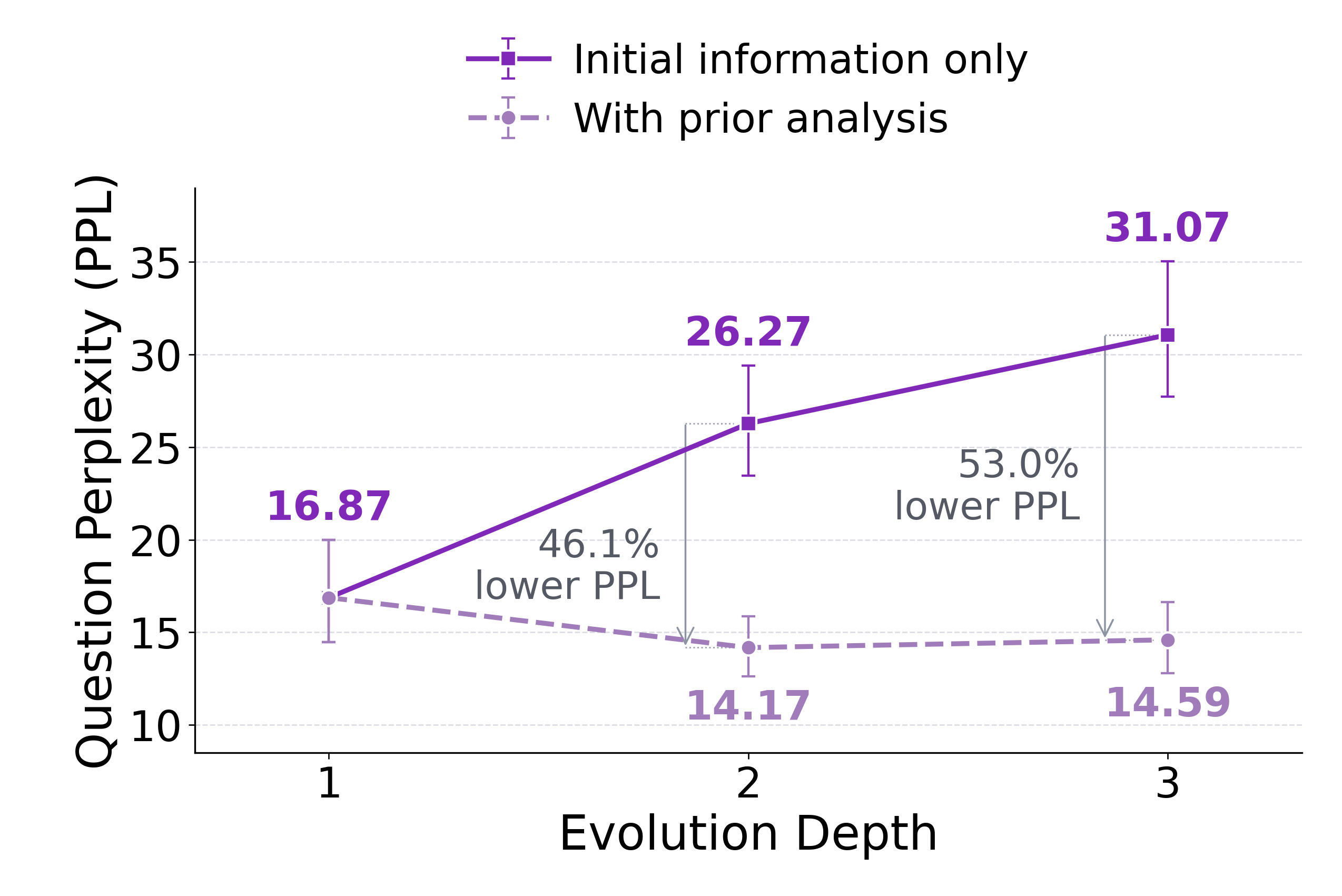}
\caption{Question perplexity with initial workspace information alone or with accumulated findings.}
\label{fig:perplexity}
\end{wrapfigure}
We use Qwen3.6-27B to measure the perplexity of each proposed question given either the initial workspace alone or the initial workspace together with findings from earlier analyses.
Figure~\ref{fig:perplexity} shows that mean perplexity given only the initial workspace rises from 16.87 to 31.07, whereas including prior findings reduces it by 46.1\% and 53.0\% in rounds 2 and 3, respectively.
These observations suggest that continued analysis gives rise to questions that are less apparent from the initial workspace, while accumulated findings provide useful context for formulating them.

Motivated by this observation, DataPraxis evolves a workspace by updating its state with verified results from each analysis.
A workspace state captures the data, scripts, and analytical results available to an agent within its execution environment at a given stage.
Each update carries forward prior analytical assets and incorporates new findings, providing the basis for subsequent analysis.
For each evolution path, DataPraxis traces the accumulated findings and their dependencies backward to synthesize one task:
\begin{equation}
X_0 \xrightarrow{\text{Evolve}} X_1
\xrightarrow{\text{Evolve}} \cdots
\xrightarrow{\text{Evolve}} X_m
\xrightarrow{\text{Synthesis}} (q,V),
\label{eq:workspace-evolution}
\end{equation}
where $m$ is the number of evolution steps and $X_1,\ldots,X_m$ are the successive workspace states.
The $q$ requires a single agent to reconstruct the analysis from $X_0$ within one session, without access to scripts, intermediate artifacts, or conclusions produced during evolution.

\WFclear
\section{Method}
\label{sec:method}
\subsection{Workspace Initialization and Profiling}
\label{sec:initialization}
As shown in Figure~\ref{fig:framework}(a), we collect raw datasets containing files in diverse formats from Kaggle, DataWorld, and GitHub repositories to construct initial workspace.
Filtering by relational structure and size yields 500 data collections spanning 15 domains, including business, government, human resources, and others.
Each collection is paired with a shared execution environment $\mathcal E$, including dependencies such as pandas and NumPy, to form a seed workspace $X_0$.

A common approach to task synthesis uses predefined task categories as seeds to guide generation~\citep{qiao2026datamind,zhang2025deepanalyze}.
However, a fixed taxonomy limits the range of questions considered, while model preferences may further concentrate generation on familiar categories.
DataPraxis therefore derives seeds from each workspace's data and domain properties.
A Profiler Agent analyzes $\mathcal D$ to construct a \emph{Workspace Profile}, capturing relationships among tables, data pitfalls, and field semantics, alongside relevant analytical methods and common domain questions.
This profile connects the available data with domain problems, guiding the generation of diverse analytical objectives suited to each workspace.

\subsection{Task Synthesis via Verified Workspace Evolution}
\label{sec:evolution}

Building on the workspace profile, DataPraxis constructs task instructions through progressive workspace evolution followed by backward synthesis.
As shown in Figure~\ref{fig:framework}(b), evolution strategy selection determines which states to inherit and how to advance the analysis.
Verified workspace evolution then updates these states through execution and verification, providing a basis for subsequent objectives.
Finally, backward task synthesis converts the accumulated analytical dependencies into task instructions, withholding intermediate support to require independent completion.

\subsubsection{Evolution Strategy}

Starting from a workspace with no predefined task, DataPraxis uses a director agent to decide where analysis should proceed.
The director considers the workspace profile and findings verified during earlier evolution steps to select eligible source workspaces.
It proposes an analytical question grounded in the available data and distinct from earlier analyses.
When extending an existing workspace, the director identifies the specific prior finding on which the new question depends.
It then selects an evolution mode according to how the question draws on the chosen workspaces.
Table~\ref{tab:operators} summarizes the four modes.
\begin{table}[htbp]
\centering
\caption{Workspace evolution modes. The mapping $s \to t$ denotes $s$ source workspaces and $t$ analytical directions, with $k \geq 2$.}
\label{tab:operators}
\begin{tabular}{lcl}
  \toprule
  Mode & Sources $\to$ Tasks & Description \\
  \midrule
  \texttt{deepen} & $1 \to 1$ &
  Extend existing findings through further analysis. \\
  \texttt{search} & $1 \to 1$ &
  Retrieve and verify relevant external materials. \\
  \texttt{decompose} & $1 \to k$ &
  Split a compound objective into independent subtasks. \\
  \texttt{compose} & $k \to 1$ &
  Merge verified findings from complementary workspaces. \\
  \bottomrule
\end{tabular}
\end{table}

Both \texttt{deepen} and \texttt{search} start from a single source workspace.
The former evolves the workspace by investigating existing findings in greater depth, while the latter expands its contents by retrieving and filtering relevant external materials.
The other two modes support workspace branching and merging.
Starting from a single workspace, \texttt{decompose} divides a compound objective into complementary subtasks, each driving the evolution of an isolated workspace.
Under \texttt{compose}, complementary verified workspaces are merged after conflicts among their contents have been resolved.
The merged workspace then evolves through an integrated analysis that requires findings from multiple sources.
Together, these modes produce diverse workspace evolution paths, enabling the synthesis of tasks with different analytical dependencies.

\subsubsection{Verified Workspace Evolution}

Each evolution step builds on inherited workspace states through workspace preparation, analytical modification, and verification, coordinated by four agents: director, proposer, executor, and validator.
The director uses the workspace profile and previously verified findings to identify an analytical question that is grounded in the data and distinct from earlier analyses.
It selects the source workspaces and an evolution mode based on what the question requires.
When extending an existing workspace, the director identifies the specific prior finding on which the new analysis depends.
The proposer then prepares the inherited workspace state by organizing its data and analytical assets.
When combining multiple sources, it removes duplicate materials and resolves file conflicts.
The proposer formulates a local analytical objective and explicit output requirements from the prepared state.
The executor updates the workspace to address this objective through reasoning and tool use.
It incorporates relevant materials, creates or revises scripts, and produces artifacts and findings.

Before this state can support further evolution, the validator reviews the changes in the workspace.
It checks whether data dependencies, metric definitions, and analytical methods are appropriate.
It also assesses whether evidence supports the conclusions and whether scripts reproduce the reported results.
If errors are found, the executor revises the affected workspace contents and submits the changes for another review.
This cycle continues until no errors are identified or the review limit is reached.
Updates that pass verification are retained as new workspace states, while those with unresolved errors are discarded.
The accepted states can then serve as sources for another evolution step, where their findings inform new objectives and their analytical assets support execution.
Repeating this procedure produces multiple evolution paths, preserving verified findings, supporting materials, and their dependencies for backward task synthesis and subsequent collection of evolution evidence.

\subsubsection{Backward Task Synthesis}
\label{sec:assembly}

Backward synthesis converts verified analyses from workspace evolution into standalone tasks and collects evolution evidence for subsequent dual-evidence verification.
DataPraxis first selects a target evolved workspace and uses its findings to define the task's objective.
It traces the evolution path backward to the initial workspace $X_0$, converting dependencies on prior findings into explicit task requirements.
For example, \texttt{search} nodes yield retrieval instructions, while computations at \texttt{deepen} nodes become prerequisite analysis steps.
The task instruction $q$ combines these requirements and specifies a deliverable format suited to the domain, such as JSON, Markdown, or a spreadsheet.
During this backward traversal, DataPraxis collects evolution evidence: the verified findings and supporting materials accumulated along the evolution path. This evidence is passed to the rubric generator and combined with solution evidence from independent reference trajectories to ground task-specific evaluation criteria.
Given $q$ and $X_0$, the solver independently plans the full workflow, reconstructs the supporting evidence, and produces the requested deliverable within one session.
This consolidates analyses previously distributed across multiple agents into one task, placing greater demands on independent planning and sustained reasoning.
Appendix~\ref{app:workspace-evolution-case} illustrates how workspace evolution and composition lead to a standalone task through a hotel pricing example.

\subsection{Dual-evidence Verification and Rejection Sampling}
\label{sec:verification}

Backward synthesis turns verified analyses into task instructions, but evidence from workspace evolution alone does not directly establish criteria for evaluating independently produced deliverables: a solver may follow a different analytical path. 
DataPraxis therefore uses dual-evidence verification to construct and validate task-specific evaluation rubrics from two evidence sources, as shown in Figure~\ref{fig:framework}(c). Evolution evidence consists of verified findings and supporting materials collected along the workspace evolution path during backward synthesis. Solution evidence is extracted from reference trajectories produced by two models that independently solve the synthesized task from the initial workspace $X_0$ in a sandbox. The rubric generator combines both sources to define criteria for conclusions, analytical methods, and delivery requirements. The rubric validator then checks whether each rubric is grounded in the data, assessable from the final deliverable, and explicit about what constitutes a satisfactory result. Rubrics that fail these checks are revised and revalidated, or removed if they remain invalid. 
Appendix~\ref{app:rubric-synthesis} presents a case study of rubric synthesis and validation.

DataPraxis packages the synthesized instruction, initial workspace, execution environment, and validated rubrics into a Harbor task.
We uses the rubrics to select teacher trajectories through rejection sampling. It assesses the final deliverables of candidate teacher trajectories against the rubrics constructed and validated using evolution evidence and solution evidence, retaining only runs that pass task-level verification. The final training dataset contains the accepted trajectories, any continuations after context compaction, and subagent trajectories from those runs.

%% file: sections/experiments_v2.tex
\section{Experiments}
\subsection{Experimental Setup}

\paragraph{Models and harness.}
We evaluate whether DataPraxis improves data analysis agents built on Qwen3.6-27B~\citep{qwen2026qwen3627b} and Qwen3.6-35B-A3B~\citep{qwen2026qwen3635ba3b}.
We use Claude Code~\citep{anthropicClaudeCode} as the harness for both trajectory collection and evaluation.
The models trained on DataPraxis-5K are denoted DataPraxis-27B and DataPraxis-35B-A3B, respectively.

\paragraph{Benchmarks.}
We evaluate the effectiveness of DataPraxis training data on three benchmarks:
(1) DABStep~\citep{egg2025dabstep} measures data analysis reasoning across multiple steps;
(2) JobBench~\citep{li2026jobbench} assesses agents' ability to analyze heterogeneous data and produce professional deliverables across occupations;
(3) RideBench is a benchmark developed internally by a company for business analysis in ride-hailing scenarios, with construction details in Appendix~\ref{app:ridebench}.
We conduct all benchmark evaluations in Harbor sandboxes and use Gemini-3.7-Flash as the judge model.

\paragraph{Baselines.}
We compare DataPraxis-27B and DataPraxis-35B-A3B against 7 leading frontier models from the Claude~\citep{anthropic2026opus5}, Kimi~\citep{moonshot2026kimik3}, GLM~\citep{zai2026glm52}, DeepSeek~\citep{deepseek2026v4}, GPT~\citep{openai2026gpt56}, and Gemini~\citep{doshi2026gemini36,doshi2026gemini37} families.
For each backbone, we also compare against the original model and a counterpart trained on 5K DataMind examples~\citep{qiao2026datamind}.
To match our agent framework and training set size, we regenerate the DataMind solution trajectories using GLM-5.2 in Claude Code.

\paragraph{Implementation and training.}
We use GLM-5.2 as the backbone model.
For dual-evidence verification, DeepSeek-V4-Pro and Kimi-K3 provide reference trajectories.
We construct a training set of 5K trajectories, including continuations after context compaction and subagent trajectories.
The agent prompts in DataPraxis are provided in Appendix~\ref{app:agent-prompts}.
We use LlamaFactory~\citep{zheng-etal-2024-llamafactory} for supervised fine-tuning.
We train the base models for 5 epochs, using a learning rate of $1\times 10^{-6}$, a batch size of 32, and a maximum sequence length of 128K tokens.

\input{sections/table_main}
\subsection{Main Results}

Table~\ref{tab:main} reports consistent improvements across both backbones and all three benchmarks after training on DataPraxis.
The average benchmark score rises from 48.36 to 60.39 for Qwen3.6-27B and from 42.43 to 55.19 for Qwen3.6-35B-A3B.
DABStep Hard shows particularly large gains in pass rate, at 23.28 and 19.31 percentage points, respectively, suggesting improved reasoning across multiple analytical steps.
For the challenging workplace tasks in JobBench, the corresponding score increases are 13.40 and 9.70 points.
RideBench scores reach 74.16 and 71.91, up by 3.37 and 12.36 points, extending the observed benefits to practical business analysis.

A comparison with DataMind further examines the effectiveness of the synthesis approach at a fixed training set size of 5K examples.
Using DataPraxis yields average benchmark scores higher by 9.14 points for Qwen3.6-27B and 13.66 points for Qwen3.6-35B-A3B.
These advantages demonstrate the value of workspace evolution for synthesizing training data for complex analytical tasks.

With just 5K training examples, both DataPraxis models also achieve scores close to or above those of several frontier models.
DataPraxis-27B is comparable to GLM-5.2 in both average benchmark score and JobBench score, while exceeding DeepSeek-V4-Pro, GPT-5.6-Terra, and Gemini-3.6-Flash in average score.
For DataPraxis-35B-A3B, the DABStep score is close to those of DeepSeek-V4-Pro.
Taken together, these comparisons demonstrate the potential of synthesized training data to make smaller models competitive with frontier models on complex analytical tasks.

\raggedbottom
\subsection{Ablation Studies}

\paragraph{Training data volume.}
Table~\ref{tab:ablations} shows that increasing the training set from 2K to 5K examples improves the DABStep score from 48.89\% to 55.11\% and JobBench from 45.50 to
51.90.
From 4K to 5K examples, DABStep gains 0.66 percentage points, while JobBench improves by a further 3.10 points.
DABStep emphasizes prerequisite computations within complex analyses, whereas JobBench requires agents to complete complex tasks and produce deliverables.
The continued improvement on JobBench suggests that additional training data remain valuable for learning to complete such tasks, even as gains on the computational subtasks become smaller.

\input{sections/table_ablations_v2}

\paragraph{Evolution depth.}
We define task evolution depth as the number of evolution steps along the longest dependency chain from the initial workspace to the workspace selected for synthesis.
To examine how depth affects task difficulty, we use Qwen3.6-35B-A3B to solve synthesized tasks independently.
Figure~\ref{fig:depth-difficulty} shows that the task pass rate decreases from 35.60\% at depth 1 to 22.90\% at depth 4, while the mean number of turns involving tool calls increases from 17.8 to 30.1.
These trends indicate that deeper tasks require more tool interaction and are harder for the model to solve.
We further examine how tasks synthesized at different evolution depths contribute to model performance.
Table~\ref{tab:depth} compares two Qwen3.6-27B models, each trained on 2K tasks.
The shallow setting uses tasks synthesized at workspace evolution depths 1--2, while the deep setting uses tasks from depths 3--5.
Compared with the shallow setting, the deep setting improves the JobBench score by 2.30 points, suggesting that the greater demands of deeper workspace evolution can benefit learning to complete complex tasks.
The shallow setting, in turn, achieves a DABStep score 1.78 percentage points higher than the deep setting, indicating the value of earlier evolution stages for learning prerequisite analytical computations.
These complementary benefits motivate selecting tasks from both shorter and longer workspace evolution paths to develop both capabilities.

\begin{figure}[htb]
\centering
\includegraphics[width=0.9\linewidth]{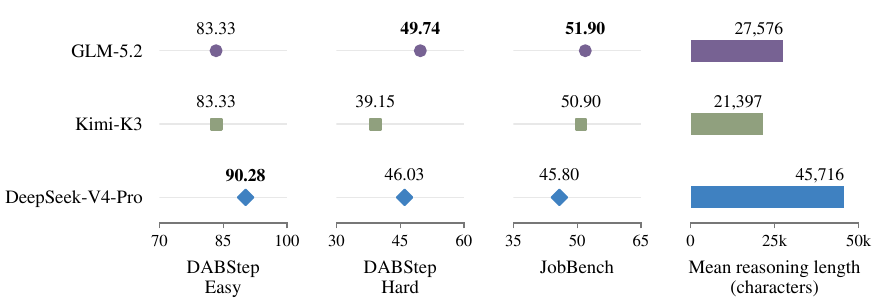}\par
\caption{Teacher ablation for Qwen3.6-27B using 5K training trajectories per teacher. Benchmark values denote student scores, with the best in bold.}
\label{fig:teacher}
\end{figure}

\paragraph{Dual-evidence verification.}
With the training set size fixed at 5K examples, removing dual-evidence verification on trajectories reduces DABStep from 55.11\% to 50.00\% and JobBench from 51.90 to 47.50.
Sampled trajectories from teacher model may contain undesirable behavioral patterns or analyses that lack sufficient evidential support.
Dual-evidence verification excludes trajectories whose deliverables fail the rubrics, helping remove unreliable solutions.

\paragraph{Trajectory teacher.}
Figure~\ref{fig:teacher} compares student performance and the mean reasoning length of training trajectories from three teachers.
Kimi-K3 has the highest average teacher score in Table~\ref{tab:main}, yet students trained on GLM-5.2 trajectories score 10.59 percentage points higher on DABStep Hard.
Compared with DeepSeek-V4-Pro, GLM-5.2 yields student scores 3.71 percentage points higher on DABStep Hard and 6.10 points higher on JobBench.
DeepSeek-V4-Pro yields the highest student score on DABStep Easy, at 90.28\%.
Teacher performance alone does not predict how much the student benefits from its trajectories.
Kimi-K3 has the lowest mean reasoning length and may leave intermediate steps implicit despite its stronger reasoning.
DeepSeek-V4-Pro averages 45,716 characters, but its longer reasoning does not yield better performance.
GLM-5.2 offers a more suitable balance of reasoning quality and explicit intermediate steps for these tasks.

\subsection{Diversity of Analytical Behaviors}
\label{sec:characterization}

\begin{figure}[htbp]
\centering
\begin{minipage}[b]{0.47\linewidth}
\centering
\includegraphics[width=\linewidth,trim={14.73bp 89.35bp 329.25bp 86.73bp},clip]{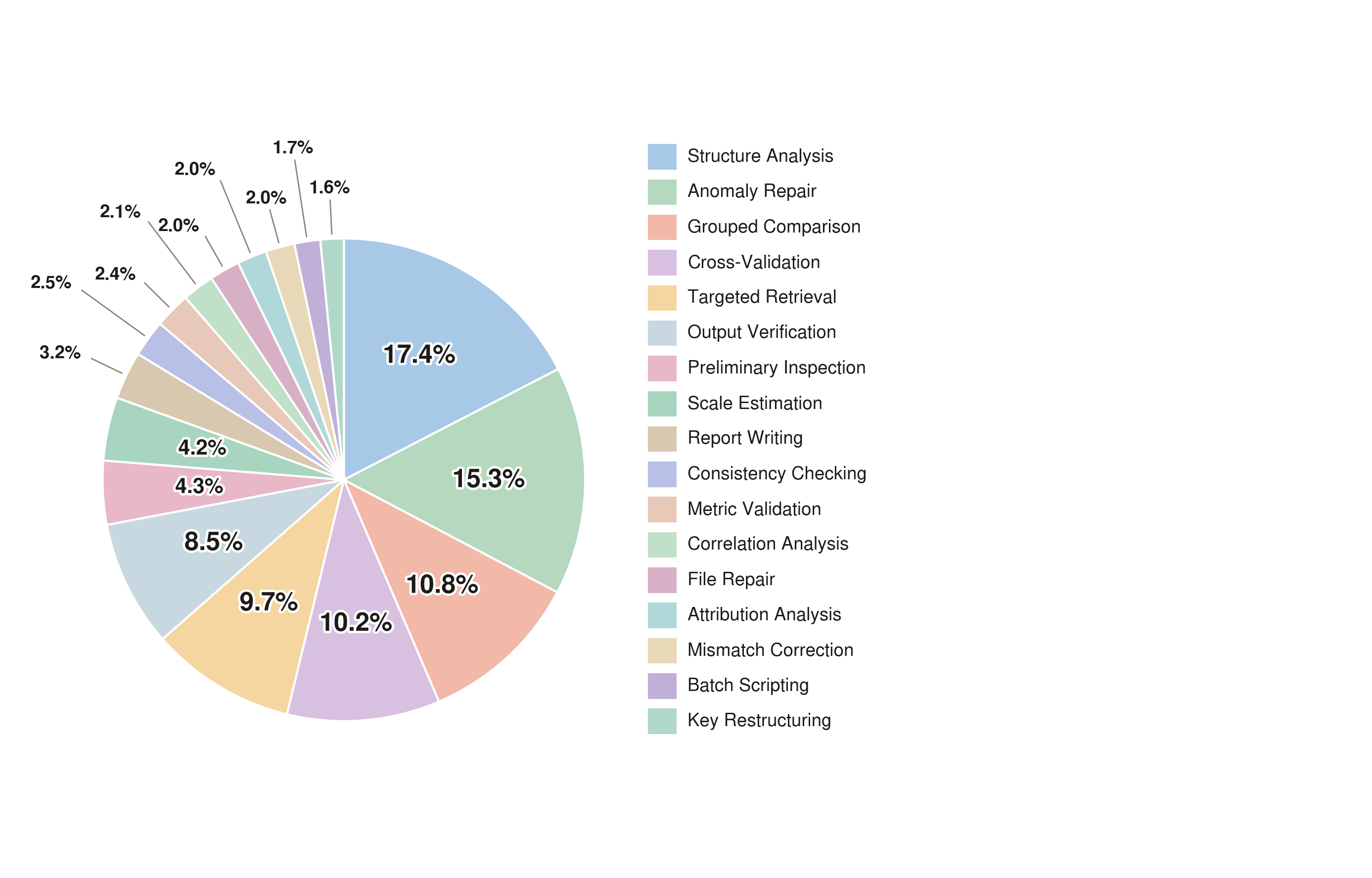}
\par (a)
\end{minipage}\hfill
\begin{minipage}[b]{0.47\linewidth}
\centering
\includegraphics[width=\linewidth]{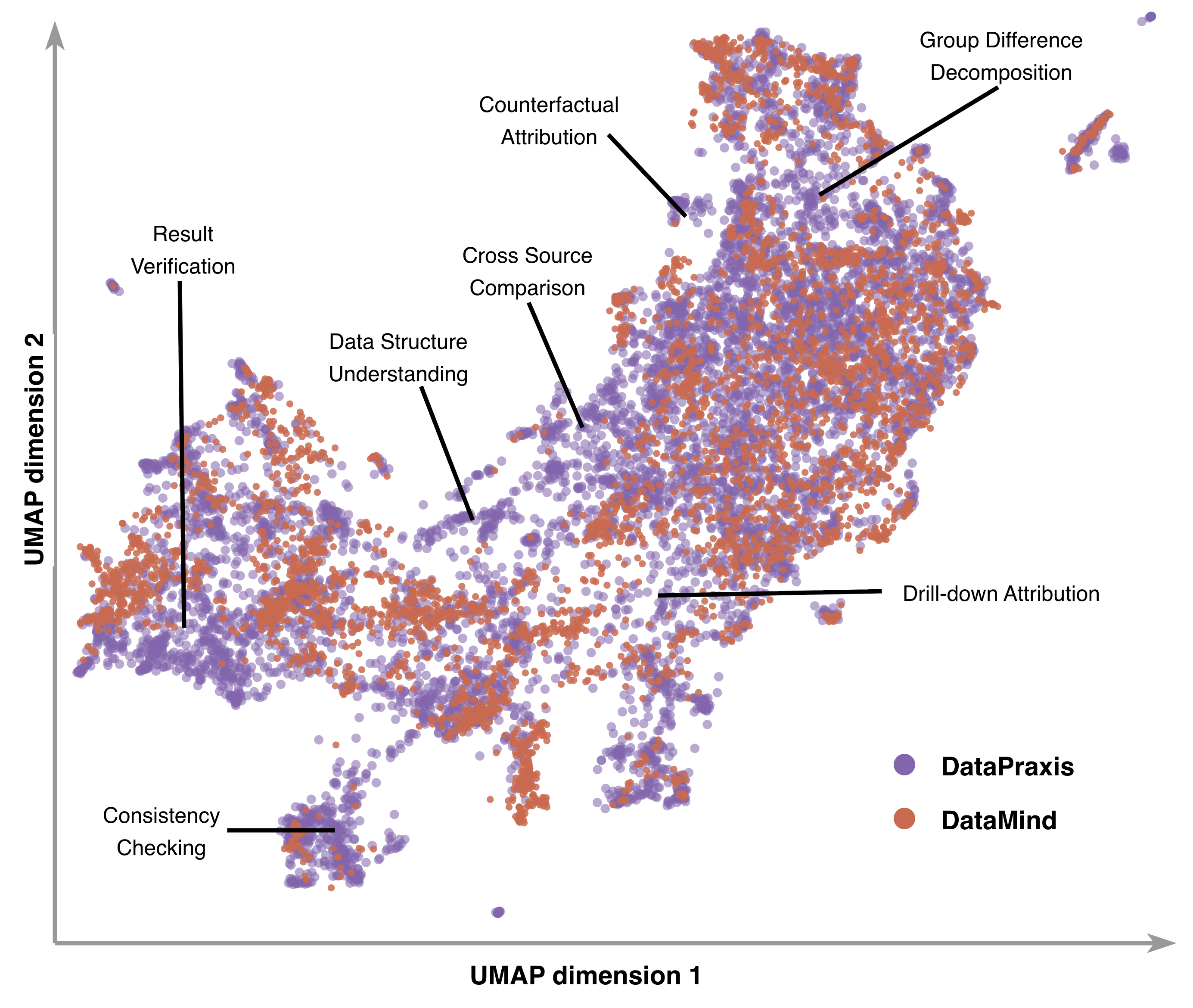}
\par (b)
\end{minipage}
\caption{Analytical diversity.
(a) Action shares across the top-level behavioral clusters derived from DataPraxis-5k.
(b) UMAP projection of action representations from DataPraxis and DataMind.}
\label{fig:analysis-panels}
\end{figure}

By varying source workspaces and evolution strategies, DataPraxis constructs tasks with different analytical requirements.
We characterize the behavior diversity in training trajectories using a hierarchical clustering procedure inspired by Clio~\citep{tamkin2024clio}.
Each action represents a coherent analytical intent.
We embed individual actions with Qwen3-4B-Embedding~\citep{zhang2025qwen3embedding}, cluster their representations, and use GLM-5.2 to name the clusters.
In each subsequent round, the names assigned in the preceding round are embedded, clustered, and named in the same manner.
Three rounds of this procedure produce a behavioral hierarchy without predefined categories.

Figure~\ref{fig:analysis-panels}(a) shows that DataPraxis trajectories span 17 top-level behavioral clusters, indicating diversity across forms of analytical work.
Clusters for structure analysis and anomaly repair appear alongside those for grouped comparison and output verification, with further behaviors represented by the remaining clusters.
Figure~\ref{fig:analysis-panels}(b) projects the action representations used in the initial clustering round into a common UMAP~\citep{mcinnes2018umap} space.
Its annotations describe the analytical behaviors captured by selected clusters.
Compared with DataMind, DataPraxis actions occupy a broader range of behavioral regions in the projection, providing qualitative evidence that its trajectories
encompass more varied analytical work.

%% file: sections/table_main.tex
  \begin{table}[tbp]
    \centering
    \caption{Performance on data analysis benchmarks. Scores marked with $\dagger$ are taken from the DABStep leaderboard.}
    \label{tab:main}
    \begin{tabular}{p{0.26\linewidth}rrrrrr}
    \toprule
    \multirow{2}{*}{Model / data}
      & \multicolumn{3}{c}{DABStep}
      & \multirow{2}{*}{JobBench}
      & \multirow{2}{*}{RideBench}
      & \multirow{2}{*}{Avg} \\
    \cmidrule(lr){2-4}
      & Easy & Hard & Overall & & & \\
    \midrule
    Claude Opus 5
      & 87.50$^{\dagger}$ & 48.68$^{\dagger}$ & 54.89 & 69.50 & 82.02 & 68.80 \\
    Kimi-K3 & 84.72 & 69.58 & 72.00 & 57.20 & 79.78 & 69.66 \\
    GLM-5.2 & 87.50 & 49.47 & 55.55 & 51.20 & 78.65 & 61.80 \\
    DeepSeek-V4-Pro & 84.72 & 44.71 & 51.11 & 49.20 & 75.28 & 58.53 \\
    GPT-5.6-Terra & 83.33 & 45.50 & 51.55 & 33.80 & 68.54 & 51.30 \\
    Gemini-3.7-Flash & 90.28 & 53.70 & 59.55 & 57.90 & 77.53 & 64.99 \\
    Gemini-3.6-Flash & 91.67 & 41.01 & 49.12 & 43.70 & 79.78 & 57.53 \\
    \midrule
    \multicolumn{7}{l}{\textbf{Qwen3.6-27B}} \\
    Base & 84.72 & 26.46 & 35.78 & 38.50 & 70.79 & 48.36 \\
    DataMind-5K & 86.11 & 38.36 & 46.19 & 37.90 & 69.66 & 51.25 \\
    \rowcolor{datapraxisblue}
    DataPraxis-5K & 83.33 & 49.74 & 55.11 & 51.90 & 74.16 & 60.39 \\
    \midrule
    \multicolumn{7}{l}{\textbf{Qwen3.6-35B-A3B}} \\
    Base & 81.94 & 27.78 & 36.45 & 31.30 & 59.55 & 42.43 \\
    DataMind-5K & 84.72 & 29.63 & 38.44 & 26.60 & 59.55 & 41.53 \\
    \rowcolor{datapraxisblue}
    DataPraxis-5K & 81.94 & 47.09 & 52.67 & 41.00 & 71.91 & 55.19 \\
    \bottomrule
    \end{tabular}
  \end{table}

%% file: sections/table_ablations_v2.tex
\newsavebox{\ablationpanelbox}
\begin{figure}[tbp]
\sbox{\ablationpanelbox}{%
\begin{minipage}[t]{0.50\linewidth}
\vspace{0pt}
\centering
\captionof{table}{Ablation results for Qwen3.6-27B. 
DataPraxis-5K uses 5K training examples with verification. DABStep reports overall
scores.}
\label{tab:ablations}
\label{tab:volume}
\label{tab:depth}
\label{tab:verify}
\begin{tabular}{@{}lrr@{}}
\toprule
Setting & DABStep & JobBench \\
\midrule
DataPraxis-5K & \textbf{55.11} & \textbf{51.90} \\
\midrule
DataPraxis-2K & 48.89 & 45.50 \\
DataPraxis-4K & 54.45 & 48.80 \\
\midrule
Shallow-2K & 53.11 & 46.50 \\
Deep-2K & 51.33 & 48.80 \\
\midrule
\shortstack[l]{w/o dual-evidence \\ verification} & 50.00 & 47.50 \\
\bottomrule
\end{tabular}
\end{minipage}%
}
\noindent\usebox{\ablationpanelbox}\hfill
\begin{minipage}[t]{0.46\linewidth}
\vspace{0pt}
\centering
\vspace*{5pt}
\includegraphics[width=\linewidth]{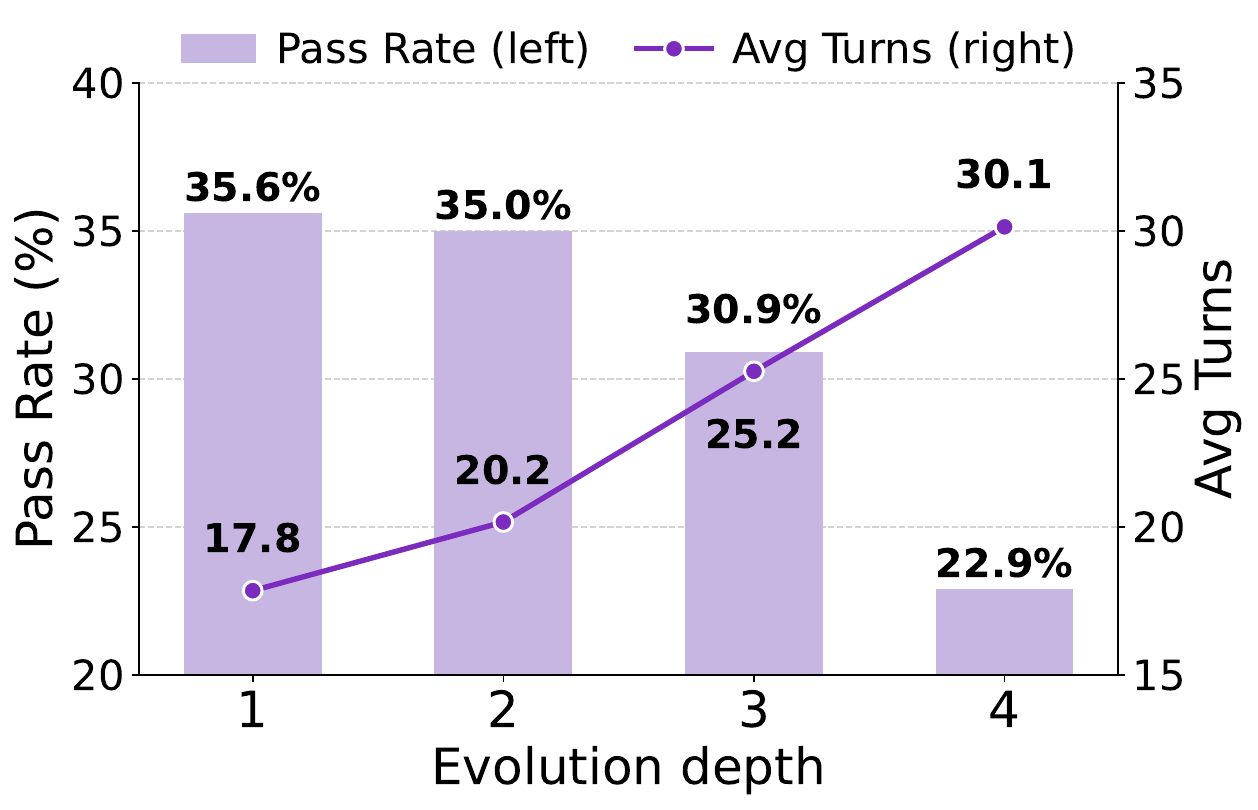}\par
\captionof{figure}{Average pass rate and tool-call turns across evolution depths.}
\label{fig:depth-difficulty}
\end{minipage}
\end{figure}

%% file: sections/related_work.tex
\section{Related Work}
\label{sec:related}

\paragraph{Task-Type and Skill Guided Task Synthesis.}
DataMind~\citep{qiao2026datamind} generates analytical questions from data metadata, fine-grained task categories, and exemplars, then increases complexity through recursive composition. DeepAnalyze~\citep{zhang2025deepanalyze} conditions interaction-task generation on sampled task types before executing and checking solution trajectories. AgenticDataBench~\citep{sun2026agenticdatabench}, SkillSynth~\citep{fan2026skillgraphs}, and SkillNet~\citep{liang2026skillnet} organize task generation through skill paths or workflows, with SkillNet additionally using exploration reports to construct tasks and evaluation components. These methods enable targeted coverage of specified tasks or skills. However, predefined categories and skill structures may constrain question discovery. Our DataPraxis continually guides task generation with verified analytical findings, uncovering deeper questions beyond these predefined structures.

\paragraph{Exploration-Based Task Synthesis.}
OS-Genesis~\citep{sun2024osgenesis} and CuES~\citep{mai2025cues} construct tasks from state transitions or interaction patterns obtained through environment exploration. AutoPlay~\citep{ramrakhya2025taskexploration} explores environment states and available functionalities before generating tasks from the resulting trajectories. WebExplorer~\citep{liu2025webexplorer} discovers interconnected facts through search and browsing before generating and iteratively refining questions. DIVE~\citep{chen2026dive} alternates tool execution, evidence accumulation, and backward question construction. Yet these methods offer limited support for reusing computations and combining analytical branches. Our DataPraxis inherits and merges verified workspaces to uncover deeper questions beyond individual branches, while retaining traceable evidence for task evaluation.

\paragraph{Compositional and Evolutionary Task Synthesis.}
AgentSynth~\citep{xie2025agentsynth} alternates the generation and execution of state-dependent subtasks before integrating them into long-horizon instructions. TaskCraft~\citep{shi2025taskcraft} introduces task dependencies through depth expansion and combines independent subtasks through width expansion. Recursive Synthesis~\citep{li2026recursivesynthesis} extends reference solutions from verified seed tasks, jointly updates instructions, environments, and verifiers, and revalidates the resulting tasks. These methods increase task complexity by composing subtasks or extending existing solutions. While adding steps can increase task complexity, it does not necessarily reveal objectives that emerge through analysis. Our DataPraxis constructs tasks backward from verified findings and their dependencies, turning these emergent objectives into analytically demanding tasks with grounded evaluation criteria.

%% file: sections/limitations.tex
\section*{Limitations}
DataPraxis relies on seed workspaces with sufficient data complexity and analytical scope to support continued exploration. 
Simple files with few variables or limited relationships may offer fewer opportunities to uncover deeper questions through workspace evolution. 
In addition, the difficulty of synthesized tasks and the quality of solution trajectories depend on the capabilities of the teacher models. 
Their ability to identify meaningful analytical directions and execute the required analyses shapes the resulting training data.

%% file: sections/ai_use_statement.tex
\subsection*{AI use statement}
We used generative AI tools to support research ideation and execution, retrieve and discover related literature, translate manuscript content, draft sections of the paper, and polish the manuscript's wording and presentation. We also used these tools to generate the synthetic dataset introduced in this work. 

Within DataPraxis, language models profile workspaces, propose and execute analytical tasks, validate workspace changes, assemble task instructions, construct and validate rubrics, and generate training trajectories. The synthesis and reference models are specified in the experimental setup. We additionally use a language model to name behavioral clusters in Section~\ref{sec:characterization}. Generated analyses and trajectories undergo the execution and verification procedures described in Section~\ref{sec:method}, with examples in Appendix~\ref{app:case-study}. The authors take responsibility for the final text, claims, and research artifacts, including content prepared with AI assistance.

%% file: sections/appendix_ridebench.tex
\section{RideBench Construction}
\label{app:ridebench}

RideBench was developed internally by a company to evaluate business analysis in ride-hailing operations.
It uses a simulator that models interactions among passenger demand, driver supply, and platform policies.
The simulator first establishes a stable operating state.
Scripted events then alter selected parameters, allowing their effects to propagate through the simulated business.
The resulting data capture changes in operational metrics under known interventions.

The benchmark covers 11 scenarios involving external events and platform strategies, including competitive pricing, subsidies, driver shortages, and extreme weather.
Some scenarios introduce a single event, while others combine interventions to support more complex attribution and strategy comparisons.
Agents receive order records and daily driver and passenger tables.
Internal simulation variables and event logs are retained separately to support reference analysis and evaluation.

The 89 tasks are organized into three difficulty levels according to the required analysis and scenario complexity.
The 56 easy tasks focus on metric computation, anomaly detection, and basic attribution.
The 23 medium tasks require more detailed attribution or inference about external events.
The 10 hard tasks emphasize operational strategy comparison and complex attribution.
Together, these tasks assess whether agents can connect observed business changes to their underlying drivers.

%% file: sections/appendix_case_studies.tex
\renewcommand{\theHfigure}{casestudy.\arabic{figure}}
\renewcommand{\theHtable}{casestudy.\arabic{table}}
\section{Case Studies}
\label{app:case-study}

\subsection{Workspace Evolution}
\label{app:workspace-evolution-case}

\begin{figure}[!htb]
\centering
\includegraphics[width=\linewidth]{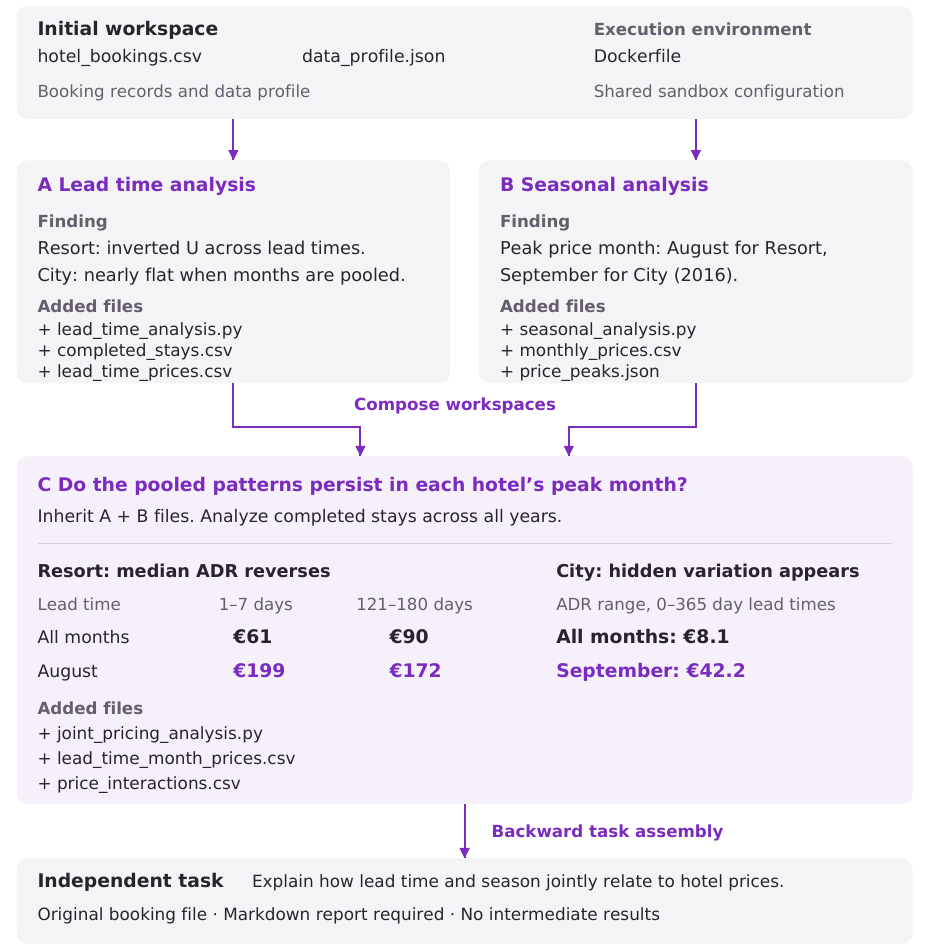}
\caption{\textbf{Workspace evolution in hotel pricing analysis.} A--C denote selected evolution nodes. Arrows show workspace inheritance and subsequent task assembly. Descriptive filenames are used for readability.}
\label{fig:workspace-evolution-case}
\end{figure}

Workspace evolution turns intermediate findings into both reusable analytical assets and grounds for subsequent questions. Figure~\ref{fig:workspace-evolution-case} illustrates this process through a hotel pricing example. Two branches characterize different dimensions of the same booking records. Their composition then tests whether the relationships found on each dimension persist when the other is taken into account.

\paragraph{Initial workspace.}
The workspace contains \texttt{hotel\_bookings.csv}, with 119,390 booking records for a City Hotel and a Resort Hotel, with arrival dates spanning July 2015 to August 2017. Available fields include the number of days between booking and arrival, the average daily room rate (ADR), hotel identity, and cancellation status. The initial snapshot also includes \texttt{data\_profile.json}. The execution environment is represented by the task package's \texttt{Dockerfile}. We follow the two pricing branches and their composition, denoted A, B, and C in the figure.

\paragraph{Node A: discovering the relationship with booking lead time.}
The first branch groups room rates by booking lead time. Among completed stays with positive room rates and stay lengths, the Resort Hotel exhibits an inverted U: median ADR rises from EUR 61 for bookings made 1--7 days before arrival to EUR 90 for bookings made 121--180 days before arrival, then declines at longer lead times. The City Hotel appears comparatively flat, with a range of only EUR 8.1 across lead time bins from 0 to 365 days. The workspace retains \texttt{lead\_time\_analysis.py}, an enriched table \texttt{completed\_stays.csv}, and grouped summaries in \texttt{lead\_time\_prices.csv}. It also establishes the cancellation filter and lead time bins used in the subsequent analysis.

\paragraph{Node B: identifying the seasonal context.}
A separate branch constructs arrival dates and examines monthly price patterns. In the complete calendar year of 2016, median ADR peaks in August for the Resort Hotel and September for the City Hotel. The same peak months remain when canceled bookings are excluded. This branch adds \texttt{seasonal\_analysis.py}, which defines the calendar mapping, together with \texttt{monthly\_prices.csv} and \texttt{price\_peaks.json}. Its findings identify specific months in which to revisit Node A's pooled relationships.

\paragraph{Node C: composing the findings into a new question.}
The composition node inherits both branches' scripts and result tables. Its recorded objective explicitly connects their findings: does the Resort Hotel's lead time premium persist in August, and does the City Hotel's apparently flat relationship persist in September? The executor reuses Node A's filtering and lead time bins together with Node B's calendar definitions to construct a joint grid of lead time and arrival month across all available years. The resulting analysis changes the interpretation of both pooled patterns. In August, the Resort Hotel's median ADR is EUR 199 for the 1--7 day bin but EUR 172 for the 121--180 day bin, reversing the corresponding pooled comparison. For the City Hotel, the range across the same 0--365 day bins expands from EUR 8.1 when months are pooled to EUR 42.2 in September. The successor workspace adds \texttt{joint\_pricing\_analysis.py}, the joint grid \texttt{lead\_time\_month\_prices.csv}, and tests in \texttt{price\_interactions.csv}, making these conditional findings available for further evolution.

\paragraph{From evolved workspace to an independent task.}
The assembled task asks a solver to analyze how booking lead time and arrival season jointly relate to room rates, assess whether their relationships can be treated separately, and compare the two hotels in a Markdown report. The solver receives the original booking file, while construction scripts, intermediate tables, and discovered conclusions are withheld. The dependencies established during evolution thus become analytical work that the solver must reconstruct. This case illustrates how composition goes beyond collecting separate results: a finding from one branch specifies where to interrogate another, producing a question and a conditional explanation that neither branch had established alone.

\subsection{Rubric Synthesis and Validation}
\label{app:rubric-synthesis}

The dual-evidence verification process turns task requirements and analytical evidence into criteria that can be checked against a deliverable. A coffee retail task illustrates how validation separates the analytical requirement from a conclusion that depends on a particular choice of analysis population. The task uses \texttt{Coffee Shop Sales.xlsx}, containing 149,116 sales rows from three stores, and requests a Markdown report on product co-occurrence and bundle recommendations. A shopping basket is defined by transaction date, time, and store.

\paragraph{From evidence to candidate criteria.}
Evolution evidence from the workspace evolution record supplies verified findings about basket composition and product associations. Solution evidence from two independent reference trajectories is organized into evidence cards linking operations, computed values, and report conclusions. The rubric generator produces ten candidate criteria covering substantive conclusions, basket construction, association measures, recommendations, and report completeness. One criterion requires distinguishing frequent co-occurrence from strong association. Its initial satisfaction rule also requires the report to conclude that Bakery and Tea have lift greater than one (Table~\ref{tab:rubric-case}).

\begin{table}[H]
\caption{Rubric refinement for coffee basket analysis. Criterion text is condensed and translated from the recorded rubric.}
\label{tab:rubric-case}
\centering
\small
\setlength{\tabcolsep}{5pt}
\renewcommand{\arraystretch}{1.16}
\begin{tabular}{@{}p{0.19\linewidth}p{0.76\linewidth}@{}}
\toprule
Stage & Selected criterion and supporting evidence \\
\midrule
Candidate & Report both co-occurrence frequency and lift. Identify Bakery + Coffee as frequent but with lift below one, and Bakery + Tea as having lift above one. \\
Validation & Bakery + Tea appears in 7,983 baskets. Its lift is 0.920 among all 116,790 baskets and 1.197 among 28,904 baskets containing multiple categories. Bakery + Coffee has lift below one under both populations (0.951 and 0.857). \\
Final criterion & Report both frequency and lift, recognize that Bakery + Coffee is frequent without a positive association, and identify at least one category pair with lift above one. The required positive pair is no longer fixed to Bakery + Tea. \\
\bottomrule
\end{tabular}
\end{table}

\paragraph{Validation establishes the scope of the expected conclusion.}
The rubric validator checks the referenced fields, the criterion's observability in the reports, and its conclusion against the workbook. For categories $A$ and $B$, lift is $n_{AB}N/(n_A n_B)$, where $N$ is the number of baskets, $n_A$ and $n_B$ count baskets containing each category, and $n_{AB}$ counts baskets containing both. All counts refer to the same basket population. Recomputing this quantity shows why the initial requirement is too specific: Bakery + Tea has lift below one over all baskets but above one after restricting to baskets containing multiple categories. The two reference reports use these different populations. The validation record therefore revises the criterion to retain the distinction between frequency and association while removing the required direction for this particular pair. The revised criterion passes the subsequent validation check.

\paragraph{Preserving a substantive scoring requirement.}
The final criterion retains a substantive test: the report must distinguish frequent co-occurrence from positive association using both frequency and lift. Listing popular combinations alone is insufficient.

\section{Model Behavior Before and After Training}
\label{app:jobbench-behavior}

We compare Qwen3.6-27B and DataPraxis-27B on the same 65 JobBench tasks across 35 occupations. We examine two observable behaviors: organizing work into explicit subgoals and checking generated outputs.

\begin{figure}[H]
\centering
\includegraphics[width=\linewidth]{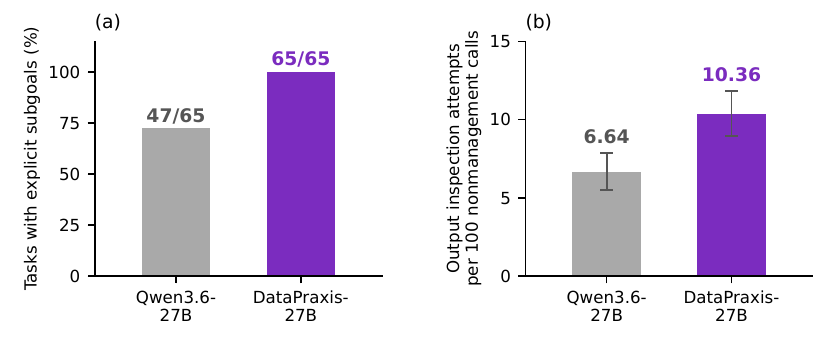}
\caption{\textbf{Task organization and output inspection on JobBench.} (a) Number of tasks with explicit subgoals, expressed as a percentage of all 65 matched tasks. Final verification goals are excluded. (b) Mean output inspection attempts per 100 nonmanagement tool calls, with 95\% bootstrap intervals over tasks. Output inspection is an observable proxy for self-verification. Both panels compare Qwen3.6-27B with DataPraxis-27B on the same task set.}
\label{fig:app-behavior-comparison}
\end{figure}

\paragraph{More explicit task organization.}
We identify explicit subgoals from successful task creation records for analysis or deliverable construction, excluding final verification goals. The number of tasks with explicit subgoals increases from 47 to 65 after training (Figure~\ref{fig:app-behavior-comparison}(a)). Among the 47 tasks with subgoals for both models, DataPraxis-27B introduces them after fewer tool calls in 44 tasks, excluding task management calls. The trained model therefore makes intermediate objectives explicit more consistently, with earlier introduction in most tasks where both models use subgoals.

\paragraph{More frequent self-verification.}
We examine self-verification through output inspection: tool calls that reopen or parse generated documents, spreadsheets, or data files to inspect their structure or contents. Checking only whether a file exists does not count. We compute inspection attempts per 100 nonmanagement tool calls for each task and average these rates equally across tasks. The mean increases from 6.64 to 10.36 per 100 tool calls, excluding task management calls (Figure~\ref{fig:app-behavior-comparison}(b)). This indicates that inspecting generated outputs occupies a larger share of tool use after training.

\subsection{Illustrative Cases of Behavioral Changes}
\label{app:workforce-case}

\begin{figure}[H]
\centering
\includegraphics[width=\linewidth]{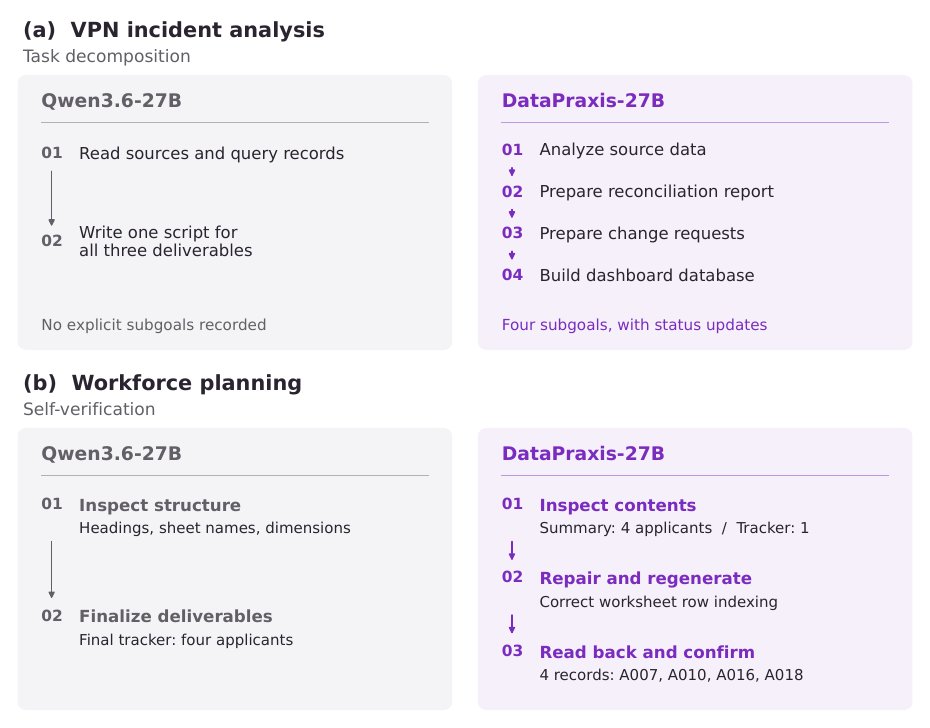}
\caption{\textbf{Paired cases illustrating the two aggregate behavior patterns.} (a) Explicit subgoals organize VPN incident analysis and its deliverables. (b) Inspecting workbook contents reveals an omission and guides its repair. Text summarizes recorded operations.}
\label{fig:app-workforce-case}
\end{figure}

We examine two tasks to illustrate the behaviors measured above. Each case compares Qwen3.6-27B and DataPraxis-27B on the same task: a VPN incident investigation illustrates explicit task decomposition, and workforce planning illustrates output inspection followed by revision (Figure~\ref{fig:app-workforce-case}).

\paragraph{Organizing a complex request into explicit subgoals.}
The VPN task requires reconciling gateway logs, incident records, and helpdesk tickets, then producing a report, a CSV containing two change requests, and a dashboard database. Qwen3.6-27B reads the sources and writes a script that generates all three deliverables, without registering explicit subgoals. After an initial inspection, DataPraxis-27B records four substantive subgoals: analyze the sources, prepare the reconciliation report, prepare the change requests, and build the dashboard. These goals retain requirements that matter to each output. For example, the change request goal specifies session and incident identifiers as evidence and requires the DNS rollback plan to account for tunnel routing discrepancies. The model then works through source analysis, report construction, change requests, and dashboard construction, updating the corresponding goal statuses as it proceeds. This makes the task's requirements explicit at the level of individual work stages, providing a concrete example of the increased subgoal coverage in Figure~\ref{fig:app-behavior-comparison}(a).

\paragraph{Using output contents to guide revision.}
In the workforce planning task, the models analyze employee turnover and screen applicants, producing a report and a workbook with a license follow-up tracker. The trajectories show how output inspection informs revision. Qwen3.6-27B inspects document headings, worksheet names, and worksheet dimensions. DataPraxis-27B additionally reads worksheet rows and summary values. Its draft summary lists four applicants requiring license follow-up, but the tracker contains only one applicant record. The subsequent edit fixes the row indexing used to populate the tracker, preventing instruction text from overwriting candidate entries. After regenerating the workbook, the model reads the tracker again and confirms all four applicant IDs. Both models ultimately deliver trackers containing these four applicants. The distinctive evidence is the trained model's observed sequence of content inspection, targeted repair, and confirmation within its own trajectory. This sequence illustrates how output inspection, the behavior measured in Figure~\ref{fig:app-behavior-comparison}(b), can lead to a concrete correction before delivery.

%% file: sections/appendix_agent_prompts.tex
\section{Agent Prompts}
\label{app:agent-prompts}
These condensed English templates retain the agents' core instructions and structured outputs. Angle brackets denote runtime inputs or values to fill in. A vertical bar separates allowed values. Use JSON null rather than a quoted string where specified, and empty lists when no entries apply. Each agent is shown once, with the Director's output selected by its operation mode.

\providecommand{\datapraxisprompt}[1]{\begingroup\setlength{\FrameSep}{6pt}\begin{framed}\small\setlength{\parskip}{3pt}\raggedright #1\end{framed}\endgroup}

\Needspace{9\baselineskip}
\subsection{Profiler}
\label{app:prompt-profiler}
\datapraxisprompt{
You are a data profiler. Inspect the initial data and research the relevant domain to produce a structured profile for subsequent exploration. No analytical objective has been assigned yet.
\medskip

{\scriptsize\ttfamily\setlength{\parskip}{0pt}
Domain context: \textless{}task\_description\textgreater{}\par
Documentation: \textless{}briefing\_files\textgreater{}\par
Search and fetch tools: \textless{}search\_tool\textgreater{}, \textless{}fetch\_tool\textgreater{}\par
Output path: \textless{}output\_path\textgreater{}\par
}

\medskip
Inspect the current workspace using scripts and suitable readers. Consult documentation before inferring field meanings. Identify artifacts, entity keys, explicit and implicit relationships, and the valuable analytical questions these relationships support. For databases, inspect schemas and basic size information without performing downstream analyses.

Research professional activities, role descriptions, analytical frameworks, and methods using the supplied search and fetch tools. Use this research to broaden the domain's work types and methods. Skip online research only if the tools are unavailable. Keep research findings in the domain profile. Do not import external numerical benchmarks or metric definitions into the workspace profile.

List atomic professional work types that can be independently performed and checked and remain meaningful for other datasets in the same domain. Separately list methods with domain concepts and their reasoning mechanisms, not generic descriptive statistics. Do not prescribe pairings between work types and methods.

Record locally observed prerequisites as usage and silent analytical pitfalls as hazard. Do not invent issues. Use Python to write only data\_profile.json to \textless{}output\_path\textgreater{}, using the following structure. The artifact note is optional.
\medskip

{\scriptsize\ttfamily\setlength{\parskip}{0pt}
\{\par
\hspace*{1.0em}"workspace\_profile": \{\par
\hspace*{2.0em}"overview": "\textless{}2-4 sentences describing the data and scale\textgreater{}",\par
\hspace*{2.0em}"artifacts": [\{"ref": "\textless{}file or table\textgreater{}", "role": "\textless{}contents\textgreater{}",\par
\hspace*{9.5em}"note": "\textless{}optional loading or usage note\textgreater{}"\}],\par
\hspace*{2.0em}"relations": ["\textless{}objects, joining keys, and analytical value\textgreater{}"],\par
\hspace*{2.0em}"observations": [\{"kind": "usage|hazard", "note": "\textless{}observation\textgreater{}"\}]\par
\hspace*{1.0em}\},\par
\hspace*{1.0em}"domain\_profile": \{\par
\hspace*{2.0em}"work\_modes": [\{"id": "\textless{}snake\_case\_id\textgreater{}", "intent": "\textless{}purpose\textgreater{}"\}],\par
\hspace*{2.0em}"domain\_methods": [\{"name": "\textless{}method\textgreater{}", "logic": "\textless{}reasoning mechanism\textgreater{}"\}]\par
\hspace*{1.0em}\}\par
\}\par
}
}

\Needspace{9\baselineskip}
\subsection{Director}
\label{app:prompt-director}
\datapraxisprompt{
You are the exploration director. Plan valuable new analytical questions from the exploration graph. Follow the supplied expansion or composition instructions.
\medskip

{\scriptsize\ttfamily\setlength{\parskip}{0pt}
Domain context and profile: \textless{}source\_task\_hint\textgreater{}\par
Candidate parents: \textless{}candidate\_list\textgreater{}\par
Exploration graph: \textless{}tree\_overview\textgreater{}\par
Depth and limits: \textless{}depth\_status\textgreater{}, \textless{}max\_depth\textgreater{}, \textless{}per\_call\_cap\textgreater{}\par
Operation instructions: \textless{}operator\_menu\textgreater{}, \textless{}required\_operator\_note\textgreater{}\par
Composition limit: \textless{}max\_compositions\textgreater{}\par
Output path: \textless{}output\_path\textgreater{}\par
}

\medskip
Read the full graph to avoid repeating existing questions. Favor substantive questions with practical or research value and support in the available data. Specify what to investigate and why, without calculation steps, formulas, expected numbers, or instructions for avoiding data pitfalls.

For expansion, choose one eligible parent and propose up to \textless{}per\_call\_cap\textgreater{} distinct directions. At the data root, derive initial questions from the profile. Otherwise identify the specific verified finding each direction depends on. A merely related topic is insufficient. Assign an allowed operator and a work\_mode to every direction. Work types are an open set, not a coverage quota. If no candidate supports a valuable new direction, set can\_expand to false, parent\_node\_id to null, and branches to an empty list.

For composition, propose up to \textless{}max\_compositions\textgreater{} natural questions, each requiring at least two verified sources. Explain their relationship and what new conclusion their combination could support. Do not merely concatenate independent questions. Keep the question in objective and the reasoning about source findings in divergence\_rationale. Return an empty list if no meaningful composition exists.

Use Python to write only director\_direction.json to \textless{}output\_path\textgreater{}. Use the expansion structure below, or the composition structure when composition is requested.
\medskip

{\scriptsize\ttfamily\setlength{\parskip}{0pt}
Expansion:\par
\{\par
\hspace*{1.0em}"can\_expand": true,\par
\hspace*{1.0em}"parent\_node\_id": "\textless{}eligible node ID, or null when stopping\textgreater{}",\par
\hspace*{1.0em}"branches": [\{"next\_focus": "\textless{}question\textgreater{}", "operator": "\textless{}allowed operator\textgreater{}",\par
\hspace*{2.0em}"depends\_on": "\textless{}parent finding, or empty at the data root\textgreater{}",\par
\hspace*{2.0em}"work\_mode": \{"id": "\textless{}work type\textgreater{}", "intent": "\textless{}purpose\textgreater{}"\},\par
\hspace*{2.0em}"insight": "\textless{}why this direction is valuable\textgreater{}"\}]\par
\}\par
Composition:\par
\{\par
\hspace*{1.0em}"operator": "composition",\par
\hspace*{1.0em}"compositions": [\{"source\_node\_ids": ["\textless{}verified ID\textgreater{}", "\textless{}verified ID\textgreater{}"],\par
\hspace*{2.0em}"relation": "\textless{}relationship\textgreater{}", "objective": "\textless{}single question\textgreater{}",\par
\hspace*{2.0em}"divergence\_rationale": "\textless{}why multiple sources are needed\textgreater{}"\}],\par
\hspace*{1.0em}"reasoning": "\textless{}rationale for the number of proposals\textgreater{}"\par
\}\par
}
}

\Needspace{9\baselineskip}
\subsection{Proposer}
\label{app:prompt-proposer}
\datapraxisprompt{
You are a goal proposer. Translate the Director's direction into a clear analytical task for the Executor. Specify what to answer and deliver, not how to calculate it.
\medskip

{\scriptsize\ttfamily\setlength{\parskip}{0pt}
Data profile: \textless{}data\_profile\textgreater{}\par
Domain context: \textless{}source\_task\_hint\textgreater{}\par
Verified predecessor findings: \textless{}history\_summary\textgreater{}\par
Direction: \textless{}director\_direction\textgreater{}\par
Operation instructions: \textless{}operator\_note\textgreater{}\par
Output path: \textless{}output\_path\textgreater{}\par
}

\medskip
Understand the core question and write a self-contained instruction describing the analytical setting, objects, question, and expected output. Keep it at a moderate scale with typically one to three central conclusions that can be reliably executed and independently checked. Do not prescribe a calculation recipe or supply the answer.

Identify the specific predecessor finding on which the new objective depends. Prepare context\_summary from history\_summary alone. Include verified findings and their source nodes so that the Executor can reuse them directly. Numerical facts must come from verified history. The Director's proposal is a direction for future work, not evidence of a result. Do not perform the analysis yourself.

Use Python to write only goal\_package.json to \textless{}output\_path\textgreater{}, using exactly this structure.
\medskip

{\scriptsize\ttfamily\setlength{\parskip}{0pt}
\{\par
\hspace*{1.0em}"current\_goal": \{\par
\hspace*{2.0em}"goal\_text": "\textless{}setting, analytical question, and expected output\textgreater{}",\par
\hspace*{2.0em}"depends\_on": "\textless{}specific predecessor finding\textgreater{}"\par
\hspace*{1.0em}\},\par
\hspace*{1.0em}"context\_summary": "\textless{}verified prior conclusions with source nodes\textgreater{}"\par
\}\par
}
}

\Needspace{9\baselineskip}
\subsection{Executor}
\label{app:prompt-executor}
\datapraxisprompt{
Complete the assigned analysis and leave reusable assets for subsequent nodes.
\medskip

{\scriptsize\ttfamily\setlength{\parskip}{0pt}
Goal: \textless{}goal\_text\textgreater{}\par
Verified prior findings: \textless{}context\_summary\textgreater{}\par
Workspace files: \textless{}workspace\_files\textgreater{}\par
Data pitfalls: \textless{}data\_traps\textgreater{}\par
Inherited scripts: \textless{}inherited\_scripts\textgreater{}\par
Inherited artifacts: \textless{}inherited\_artifacts\textgreater{}\par
Node index and output path: \textless{}round\_idx\textgreater{}, \textless{}output\_path\textgreater{}\par
}

\medskip
Reuse or extend verified scripts and intermediate artifacts under sources/. Compute from raw data when an inherited result is unavailable. Write analyze\_node\textless{}round\_idx\textgreater{}.py as a self-contained script that reproduces the key numerical evidence from persisted inputs, without manual steps or transient state. Save new intermediate artifacts under generated/.

Derive directional and quantitative claims from computed values. If findings overturn an expectation, correct the script and rerun it. Do not patch the result JSON while leaving contradictory code. State conclusion as prose with its supporting numbers inline so subsequent nodes can reuse it directly.

Write analysis\_log\_node\textless{}round\_idx\textgreater{}.md documenting the subproblems, assumptions and processing choices, calculation steps and intermediate values, changes prompted by the findings, and the evidence supporting each final claim. Include both the script and log in generated\_files, together with reusable artifacts.

Write execution\_result.json to \textless{}output\_path\textgreater{} with all fields below. Set execution\_status to success or failed. Use null for failure\_reason on success. A successful conclusion must be nonempty prose, not a structured table.
\medskip

{\scriptsize\ttfamily\setlength{\parskip}{0pt}
\{\par
\hspace*{1.0em}"execution\_status": "success|failed",\par
\hspace*{1.0em}"observations": ["\textless{}finding with supporting values\textgreater{}"],\par
\hspace*{1.0em}"methodology": "\textless{}analytical method and processing choices\textgreater{}",\par
\hspace*{1.0em}"conclusion": "\textless{}conclusion with supporting numbers inline\textgreater{}",\par
\hspace*{1.0em}"generated\_files": ["analyze\_node\textless{}round\_idx\textgreater{}.py",\par
\hspace*{2.0em}"analysis\_log\_node\textless{}round\_idx\textgreater{}.md", "generated/\textless{}artifact\textgreater{}"],\par
\hspace*{1.0em}"failure\_reason": null\par
\}\par
}
}

\Needspace{9\baselineskip}
\subsection{Validator}
\label{app:prompt-validator}
\datapraxisprompt{
You are an analysis validator. Try to falsify the main conclusion fairly, without inventing objections. Determine reliability and establish the verified conclusion.
\medskip

{\scriptsize\ttfamily\setlength{\parskip}{0pt}
Goal: \textless{}goal\_text\textgreater{}\par
Execution result: \textless{}execution\_result\_json\textgreater{}\par
Workspace files: \textless{}workspace\_files\textgreater{}\par
Output path: \textless{}output\_path\textgreater{}\par
}

\medskip
Check execution success, listed files, and substantive observations and conclusion. Failure of any of these checks means unsupported with a null ground\_truth. Do not rescue a failed execution by independently completing it. Inspect the required script and analysis log. Rerun the script to check independent reproducibility, which does not itself establish correctness.

Identify the strongest plausible objection. Trace claims to calculations and assumptions, check the appropriateness of methods, and confirm that the analysis answers the assigned question. Judge the chosen analytical perspective on its own terms.

Return supported if the main claim withstands scrutiny. Return revised only after a bounded correction to the existing computation, such as repairing a formula or filter, and rerun the corrected script. If a reliable correction requires new analysis, return unsupported. Preserve sources/ and the original execution\_result.json.

Write ground\_truth in your own words with verified numbers inline. Record unresolved concerns in issues, and inherited assumptions or intermediate artifacts in condition\_dependencies. Use Python to save validation\_result.json at \textless{}output\_path\textgreater{} with the structure below. Set ground\_truth to null for unsupported. Include zero to three issues, using major, moderate, or minor severity. For revised, revisions must identify script changes and the recomputed values.
\medskip

{\scriptsize\ttfamily\setlength{\parskip}{0pt}
\{\par
\hspace*{1.0em}"main\_claim": "supported|revised|unsupported",\par
\hspace*{1.0em}"ground\_truth": "\textless{}verified prose conclusion, or null if unsupported\textgreater{}",\par
\hspace*{1.0em}"revisions": ["\textless{}correction and recomputed evidence\textgreater{}"],\par
\hspace*{1.0em}"issues": [\{"severity": "major|moderate|minor",\par
\hspace*{7.0em}"description": "\textless{}specific issue\textgreater{}"\}],\par
\hspace*{1.0em}"condition\_dependencies": ["\textless{}inherited assumption or artifact\textgreater{}"],\par
\hspace*{1.0em}"summary": "\textless{}what was checked and whether the conclusion holds\textgreater{}"\par
\}\par
}
}

\Needspace{9\baselineskip}
\subsection{Verifier Generation}
\label{app:prompt-verifier-generation}
\datapraxisprompt{
You are a read-only rubric generation agent. Derive criteria from the task, data, solving evidence, and candidate reference conclusions. Reference candidates may be wrong and must not be treated as guaranteed truth.
\medskip

{\scriptsize\ttfamily\setlength{\parskip}{0pt}
Instruction: \textless{}instruction\textgreater{}\par
Data files: \textless{}data\_file\_paths\textgreater{}\par
Evidence cards: \textless{}evidence\_cards\textgreater{}\par
Reference candidates: \textless{}gt\_candidate\_rubrics\textgreater{}\par
}

\medskip
Use Read, Bash, Grep, and Glob without modifying files. Read only supplied paths. Inspect the task, data schema, evidence cards, and reference candidates. Induce four to seven evaluation dimensions covering main conclusions, analytical methods, and presentation, with greatest importance given to the main conclusions. Assess presentation from extracted text, not rendered pixels.

Generate one to three criteria per dimension and at most ten overall. Each criterion must test one observable concern with a justified expected outcome. Drop correctness criteria whose expected outcome you cannot establish. Do not accept opposite answers, mere self-consistency, or answer presence as correctness. Allow genuinely open answers justified by the task and data, and completeness checks for explicitly required presentation.

Use filter for fixed primary conclusions and preference for method or presentation. Label provenance honestly as gt or generated. A downstream step caps gt criteria at 40 percent, dropping those with lowest confidence if exceeded. Deduplicate across all sources and record removals.

Return exactly one minified JSON object in the final reply, with the following fields. Do not use Markdown fences or additional prose. Tool output is not the final reply. Use null for source\_field when no reference field applies.
\medskip

{\scriptsize\ttfamily\setlength{\parskip}{0pt}
\{\par
\hspace*{1.0em}"rubric\_id": "\textless{}ID\textgreater{}", "instruction\_key": "\textless{}key\textgreater{}",\par
\hspace*{1.0em}"evaluation\_dimensions": ["\textless{}dimension\textgreater{}"],\par
\hspace*{1.0em}"criteria": [\{\par
\hspace*{2.0em}"criterion\_id": "F01", "type": "filter|preference",\par
\hspace*{2.0em}"dimension": "\textless{}one evaluation dimension\textgreater{}",\par
\hspace*{2.0em}"criterion": "\textless{}declarative expected outcome\textgreater{}",\par
\hspace*{2.0em}"observable": "\textless{}evidence to inspect\textgreater{}",\par
\hspace*{2.0em}"satisfy": "\textless{}conditions for satisfying the criterion\textgreater{}",\par
\hspace*{2.0em}"violate": "\textless{}conditions for violating the criterion\textgreater{}",\par
\hspace*{2.0em}"confidence": 0.0, "source": "gt|generated",\par
\hspace*{2.0em}"source\_field": "\textless{}ground\_truth field, or null\textgreater{}",\par
\hspace*{2.0em}"source\_note": "\textless{}reference conclusion or independently derived angle\textgreater{}"\par
\hspace*{1.0em}\}],\par
\hspace*{1.0em}"rubric\_warnings": [], "dedup\_notes": ["\textless{}removed duplicate\textgreater{}"]\par
\}\par
}
}

\Needspace{9\baselineskip}
\subsection{Verifier Validation}
\label{app:prompt-verifier-validation}
\datapraxisprompt{
You are a read-only rubric validation agent. Validate one criterion against the task, input data, solving evidence, and delivered output text.
\medskip

{\scriptsize\ttfamily\setlength{\parskip}{0pt}
Criterion and task: \textless{}criterion\textgreater{}, \textless{}instruction\textgreater{}\par
Data files: \textless{}data\_file\_paths\textgreater{}\par
Evidence: \textless{}evidence\_cards\textgreater{}, \textless{}output\_files\_by\_trajectory\textgreater{}\par
}

\medskip
Use Read, Bash, Grep, and Glob without modifying files. Read only supplied paths and never read ground\_truth files. Recompute from input data to establish correctness, independently of agreement among trajectories.

Check four angles. Grounding checks whether referenced fields and concepts exist. Observability checks whether supplied text and evidence reveal satisfaction or violation, excluding requirements that need rendered pixels. Formulation checks one clear declarative requirement with meaningful satisfy and violate conditions. Reject correctness clauses that accept opposite answers, mere self-consistency, or only answer presence. Allow justified open questions and explicitly required presentation completeness. Leave neutral clauses that do not exhibit these defects unchanged. Conclusion correctness checks whether the expected outcome holds, using as much recomputation as the claim requires.

Distinguish verified, unverifiable, over\_specified, and refuted outcomes. Unverifiable means the data cannot settle the claim, not that a feasible check was skipped. An arbitrary demand for exact values or rankings is over\_specified. Do not reject a criterion merely because all current trajectories satisfy it.

Leave scoring\_tip empty unless equally valid methods reach the same primary conclusion and differ only in secondary details. Pass only if grounding is supported or ambiguous, observable is observable or partially\_observable, well\_formed is well\_formed, and conclusion\_correct is verified or unverifiable. Return exactly one minified JSON object with the following fields, without Markdown or additional prose.
\medskip

{\scriptsize\ttfamily\setlength{\parskip}{0pt}
\{\par
\hspace*{1.0em}"criterion\_id": "\textless{}ID\textgreater{}",\par
\hspace*{1.0em}"validation\_angles": \{\par
\hspace*{2.0em}"grounding": \{\par
\hspace*{3.0em}"assessment": "supported|unsupported|ambiguous", "reason": "\textless{}evidence\textgreater{}",\par
\hspace*{3.0em}"data\_evidence": [\{"file": "\textless{}path\textgreater{}", "sheet": "\textless{}sheet\textgreater{}", "fields": ["\textless{}field\textgreater{}"]\}]\par
\hspace*{2.0em}\},\par
\hspace*{2.0em}"observable": \{\par
\hspace*{3.0em}"assessment": "observable|partially\_observable|unobservable", "reason": "\textless{}evidence\textgreater{}"\par
\hspace*{2.0em}\},\par
\hspace*{2.0em}"well\_formed": \{\par
\hspace*{3.0em}"assessment": "well\_formed|ill\_formed", "reason": "\textless{}evidence\textgreater{}", "issues": []\par
\hspace*{2.0em}\},\par
\hspace*{2.0em}"conclusion\_correct": \{\par
\hspace*{3.0em}"assessment": "verified|unverifiable|over\_specified|refuted",\par
\hspace*{3.0em}"reason": "\textless{}evidence\textgreater{}", "verification": "\textless{}check, finding, and criterion claim\textgreater{}"\par
\hspace*{2.0em}\}\par
\hspace*{1.0em}\},\par
\hspace*{1.0em}"overall\_pass": true, "failed\_angles": ["\textless{}failed angle\textgreater{}"],\par
\hspace*{1.0em}"scoring\_tip": "", "agent\_confidence": 0.0\par
\}\par
}
}